\documentclass{llncs}

\usepackage[english]{babel}
\usepackage{color}
\usepackage{amsmath,amsfonts,amssymb}
\usepackage{bigints}
\usepackage{graphicx}

\DeclareMathOperator*{\argmin}{arg\,min}

\title{Constructive Setting of the Density Ratio Estimation Problem and its Rigorous Solution}

\author{Vladimir Vapnik\inst{1} \and Igor Braga\inst{2} \and Rauf Izmailov\inst{3}}

\institute{ NEC Laboratories America, Princeton, NJ 08540, USA
\and
Institute of Mathematics and Computer Science \\ University of S\~ao Paulo, S\~ao Carlos, SP 13566-590, Brazil
\and
Applied Communication Sciences \\ 150 Mt Airy Road, Basking Ridge, NJ 07920, USA \\
\email{vlad@nec-labs.com, igorab@icmc.usp.br, rizmailov@appcomsci.com}
}

\begin{document}

\maketitle

\begin{abstract}
We introduce a general constructive setting of the density ratio
estimation problem as a solution of a (multidimensional) integral
equation. In this equation, not only its right hand side is known
approximately, but also the integral operator is defined
approximately. We show that this ill-posed problem has a rigorous
solution and obtain the solution in a closed form. The key element
of this solution is the novel $V$-matrix, which captures the
geometry of the observed samples. We compare our method with three
well-known previously proposed ones. Our experimental results
demonstrate the good potential of the new approach.
\end{abstract}


\section{Introduction}

The estimation of the ratio of two probability densities from a
given collection of data is a fundamental technique for solving many
applied problems, including data
adaptation~\cite{shimodaira:00:jspi,sugiyama:12:book}, conditional
probability and regression estimation~\cite{vapnik:98:book},
estimation of mutual information~\cite{suzuki:09:isit},
change-point detection~\cite{kawahara:09:sdm}, and many others.

Several approaches have been proposed and studied for the solution
of this problem (see~\cite{sugiyama:11:book},
\cite{gretton:09:book}, \cite{nguyen:07:nips}, \cite{nguyen:10:tit}
and references therein). Among them, we find the Kernel Mean Matching
(KMM) procedure~\cite{gretton:09:book}, the unconstrained
Least-Squares Importance Filtering (uLSIF)
algorithm~\cite{kanamori:09:jmlr}, and the Kullback-Leibler
Importance Estimation Procedure (KLIEP)~\cite{sugiyama:07:nips}. For
some of the proposed approaches, the convergence of the obtained
estimates to the actual solution was proven under some assumptions
about the (unknown) solution.

In this paper we introduce direct constructive methods of density
ratio estimation. As opposed to existing approaches, ours is based
on the definition of the density ratio function itself. We show that
density ratio estimation requires the solution of an ill-posed
integral equation for which not only the right hand side of the
equation is approximately defined, but so is the operator of the
equation. The solution of such equation essentially depends on our
new concept of ``$V$-matrix'', which is directly computed from
data. This type of matrix was not used in previously proposed
approaches.

The paper is organized as follows. In Section~\ref{sec:theory} and
\ref{sec:theory-ratio} we outline the necessary basic concepts of
theoretical statistics and show their relation to direct
constructive density ratio estimation. In Section~\ref{sec:norm}
through \ref{sec:gamma}, we derive the concept of $V$-matrix and
direct methods of density ratio estimation based on this concept.
Section~\ref{sec:exp} is devoted to experimental results.



\section{Statistical Foundations}
\label{sec:theory}

In order to simplify the notations in this paper, we denote a
multidimensional function $f(x^1,\ldots,x^d)$ by $f(x)$. Likewise,
we use the following notation for multidimensional integrals:
\begin{equation*}
\int_{s}^{u} {f(t) \, dg(t)} \equiv \int_{s^1}^{u^1}\text{\scriptsize{...}}\int_{s^d}^{u^d} {f(t^1,\ldots,t^d)dg(t^1,\ldots,t^d)}.
\end{equation*}

\subsection{Empirical Distribution Function}

The basic concept of theoretical statistics is the
cumulative distribution function (CDF) $F$ of a random vector $X=(X^1,\ldots,X^d)$:
\begin{equation*}
F(x^1,\ldots,x^d) = P\{(X^1 \leq x^1) \cap \ldots \cap (X^d \leq x^d)\}.
\end{equation*}

It is known that any CDF $F(x)$ is well-approximated by the
empirical cumulative distribution function (ECDF) $F_\ell(x)$
constructed on the basis of an i.i.d. sample distributed according
to $F(x)$:
$$x_1,\ldots,x_\ell~\stackrel{i.i.d}{\sim}~F(x).$$
The empirical cumulative distribution function $F_\ell(x)$ has the
form
\begin{equation}
  F_\ell(x^1,\ldots,x^d) = \frac{1}{\ell}\sum_{i=1}^{\ell}{\prod_{k=1}^{d}{\theta(x^k-x_i^k)}},
  \label{eq:edf}
\end{equation}
where $\theta(t)$ is the step-function\footnote{Step-function
$\theta(t)$ is defined as\begin{equation*}
  \theta(t) = \left\{
  \begin{array}{l l}
    1, & \quad \text{if}\ t \geq 0 \\
    0, & \quad \text{otherwise.} \\
  \end{array} \right.
\end{equation*}}.
For simplicity, we refer to the multidimensional $F_\ell(x^1,\ldots,x^d)$ from now on as
\begin{equation}\label{eq:femp}
  F_\ell(x) = \frac{1}{\ell}\sum_{i=1}^{\ell}{\theta(x-x_i)}.
\end{equation}


For $d=1$, the Dvoretzky-Kiefer-Wolfowitz inequality states that
$F_\ell(x)$ converges fast to $F(x)$, \textit{namely}, for any
natural $\ell$ and any $\epsilon>0$,
\begin{equation*}
    P\left\{\sup_{x}|{F_\ell(x) - F(x)}| > \epsilon\right\} \leq 2\,\text{exp}({-2\epsilon^{2}\ell}).
\label{eq:kolm-ineq}
\end{equation*}

For $d>1$, fast convergence of $F_\ell(x)$ to $F(x)$ also takes place:
\begin{equation}
    P\left\{\sup_{x}|{F_\ell(x) - F(x)}| > \epsilon\right\} \leq 2 \text{exp}({-c^*\epsilon^{2}\ell}),
\label{eq:vc-bound}
\end{equation}
where, according to VC theory~\cite{vapnik:98:book}, $c^*$ can be set to
\begin{equation*}
    c^* = 1 - \frac{(d-1) \ln \ell}{\epsilon^2\ell}.
\end{equation*}

\subsection{Constructive Setting of the Density Estimation Problem}

A probability density function $p(x^1,\ldots,x^d)$ (if it exists) is
defined as the derivative of the CDF:
\begin{equation}
p(x^1,\ldots,x^d) = \frac{\partial^d F(x^1,\ldots,x^d)}{\partial x^1 \ldots \partial x^d}.
\label{eq:dens}
\end{equation}

According to the definition of density function \eqref{eq:dens}, the problem of density
estimation from a given collection of data is the problem of solving the (multidimensional) integral equation
\begin{equation}
    \int_{-\infty}^{x}{p(t)\,dt} = F(x)
    \label{eq:int-dens}
\end{equation}
when the cumulative distribution $F(x)$ is unknown but an
\textit{i.i.d} sample $$x_1,\ldots,x_\ell~\sim~F(x)$$ is given.

The constructive setting of this problem is to solve equation
\eqref{eq:int-dens} using approximation (\ref{eq:femp}) instead of $F(x)$. As follows from (\ref{eq:vc-bound}), $F_\ell$
converges to $F$ uniformly with probability $1$.

\subsection{Ill-Posed Problems}

It is known that solving a linear operator equation
\begin{equation}
Af=F
\label{eq:op-equation}
\end{equation}
such as \eqref{eq:int-dens} is an ill-posed problem: small
deviations in the right hand side $F$ may lead to large deviations
of the solution $f$.

In what follows, we assume that the operator $A$ maps functions from
a normed space $E_1$ to functions in a normed space $E_2$.

In the 1960's, Tikhonov proposed the regularization method for solving ill-posed problems; it uses approximations $F_\ell$ such that
\begin{equation*}
\left\|F-F_\ell\right\|_{E_2} \rightarrow 0\ \ \text{as}\ \ \ell \rightarrow
\infty.
\end{equation*}
According to this method, in order to solve an ill-posed
problem, one has to minimize the functional
\begin{equation*}
f_\ell = \argmin_{f \in \mathcal{F}} [\ \left\|Af-F_\ell\right\|^2_{E_2} + \gamma_\ell\,\Omega(f)\ ],
\label{eq:tikh-func}
\end{equation*}
where $\gamma_\ell > 0$ is a regularization constant and $\Omega(f)$
is a regularizing functional defined in the space $E_1$; the
functional $\Omega(f)$ must possess the following properties:
\begin{itemize}
  \item $\Omega(f)$ is non-negative;
  \item there exists a $c_0< \infty$ for which the solution $f_0$ of \eqref{eq:op-equation} is in $\{\Omega(f) \leq c_0\}$;
    \item for every $c$, the set of functions $\{\Omega(f) \leq c\}$ is \emph{compact}.
\end{itemize}

It was shown for $f_\ell$ that,  if the rate of convergence of
$\gamma_\ell \rightarrow 0$ (as $\ell \rightarrow \infty$) is not
greater than the rate of convergence of
$\left\|F-F_\ell\right\|_{E_2} \rightarrow 0$, then
\begin{equation*}
\left\|f_0-f_\ell\right\|_{E_1} \rightarrow 0.
\end{equation*}

For the stochastic case, in which $F_\ell$ converges in probability to $F$,
it was shown that, if $\gamma_\ell \rightarrow 0$ as $\ell \rightarrow \infty$,
then for arbitrary $\delta,\nu > 0$ there is a $\ell>\ell_0(\delta,\nu)$ such that the
following inequality holds for $f_\ell$~\cite{vapnik:78:arc,vapnik:98:book}:
\begin{equation}
P\{\left\|f_0-f_\ell\right\|_{E_1} > \delta\} \leq P\{\left\|F-F_\ell\right\|_{E_2} > \sqrt{\gamma_\ell\nu}\}.
\label{eq:dens-ineq}
\end{equation}

In particular, for density estimation inequalities
\eqref{eq:vc-bound} and \eqref{eq:dens-ineq} imply that
\begin{equation*}
P\{\left\|p_0-f_\ell\right\|_{E_1} > \delta\} \leq c\,\text{exp}(-c^* \gamma_\ell \nu \ell).
\end{equation*}
Therefore, if
\begin{equation*}
  \gamma_\ell \xrightarrow{\ell \rightarrow \infty} 0\ \ \mbox{and}\ \ \ell\gamma_\ell \xrightarrow{\ell \rightarrow \infty} \infty,
  \label{eq:cond-dens}
\end{equation*}
the sequence $f_\ell$ converges in probability to the solution of equation \eqref{eq:int-dens}.


\section{Constructive Setting of the Density Ratio Estimation Problem}
\label{sec:theory-ratio}

In what follows, we consider the problem of estimating the ratio
$r(x)$ of two probability densities $p_1(x)$ and $p_2(x)$ (assuming
$p_2(x)
> 0$):
\begin{equation}
  r(x) = \frac{dF_1(x)/dx}{dF_2(x)/dx} = \frac{p_1(x)}{p_2(x)}.
  \label{eq:ratio}
\end{equation}

According to definition \eqref{eq:ratio}, the problem of
estimating the density ratio from data is the problem of solving the
(multidimensional) integral equation
\begin{equation}
    \int_{-\infty}^{x}{r(t)\,dF_2(t)} = F_1(x)
    \label{eq:int-ratio}
\end{equation}
when the distribution functions $F_1(x)$ and $F_2(x)$ are unknown but
samples $x_1,\ldots,x_\ell~\stackrel{i.i.d}{\sim}~F_1(x)$
and $x'_1,\ldots,x'_n~\stackrel{i.i.d}{\sim}~F_2(x)$ are given.

The constructive setting of this problem is to solve equation \eqref{eq:int-ratio} using the empirical distribution
functions
\begin{equation*}
  F_{1,\ell}(x) = \frac{1}{\ell}\sum_{i=1}^{\ell}{\theta(x-x_i)} \ \ \ \mbox{and} \ \ \ F_{2,n}(x) = \frac{1}{n}\sum_{i=1}^{n}{\theta(x-x'_i)}
\end{equation*}
instead of the actual cumulative distributions $F_1(x)$ and
$F_2(x)$.

\subsection{Stochastic Ill-Posed Problems}

Note that density ratio estimation leads to a more complicated ill-posed equation
than density estimation, since the operator $A$, which depends on $F_2(x)$,
is also defined approximately:
\begin{equation*}
(A_nr)(x) = \int_{-\infty}^{x}{r(t)\,dF_{2,n}(t)} = \frac{1}{n}\sum_{i=1}^{n}{r(x'_i)\,\theta(x-x'_i)}.
\label{eq:emp-op}
\end{equation*}

In this situation, in order to solve equation \eqref{eq:int-ratio},
we will also minimize the regularization functional
\begin{equation}
r_{\ell,n} = \argmin_{r \in \mathcal{F}} [\ \left\|A_n r-F_{1,\ell}\right\|_{E_2}^2 + \gamma_{\ell,n}\,\Omega(r)\ ].
\label{eq:tikh-func-ratio}
\end{equation}

Let the sequence of operators $A_n$ converge in probability to $A$
in the following operator norm:
\begin{equation*}
  \left\|A  - A_n\right\| = \sup_{r \in \mathcal{F}} \frac{\left\|Ar-A_n r\right\|_{E_2}}{\sqrt{\Omega(r)}}.
\end{equation*}
Then, for arbitrary $\delta,C_1,C_2 > 0$, there exists $\gamma_0 >
0$ such that for any $\gamma_{\ell,n} \leq \gamma_0$ the following
inequality holds for
$r_{\ell,n}$~\cite{vapnik:98:book,stefanyuk:86:arc}:
\begin{equation}
P\{\left\|r_0 - r_{\ell,n}\right\|_{E_1} > \delta \} \leq P\{\left\|A
- A_n\right\| > C_2 \sqrt{\gamma_{\ell,n}}\} + P\{\left\|F_1 -F_{1,\ell}\right\|_{E_2} > C_1 \sqrt{\gamma_{\ell,n}}\}.
\label{eq:ratio-ineq}
\end{equation}

It was shown \cite{vapnik:98:book} that, if the solution of the
equation \eqref{eq:int-ratio} belongs to a set $\mathcal{F}$ of
smooth functions (in fact, to a set of continuous functions with
bounded variation), then
\begin{equation}
  \left\|A  - A_n\right\| \leq \left\|F_2  - F_{2,n}\right\|_{E_2}.
  \label{eq:op-approx}
\end{equation}

For sufficiently large $\ell$ and $n$, inequalities
\eqref{eq:vc-bound}, \eqref{eq:ratio-ineq}, and \eqref{eq:op-approx}
imply that
\begin{equation*}
P\{\left\|r_0 - r_{\ell,n}\right\|_{E_1} > \delta\} \leq
c_2\,\text{exp}(-c^*_2 n \gamma_{\ell,n} C_2^2) + c_1\,
\text{exp}(-c^*_1 \ell \gamma_{\ell,n} C_1^2).
\end{equation*}

Therefore, in the set $\mathcal{F}$ of smooth functions, the
sequence $r_{\ell,n}$ converges in probability to the solution of equation \eqref{eq:int-ratio} provided that
\begin{equation}
  \gamma_{\ell,n} \xrightarrow{m \rightarrow \infty} 0\ \ \mbox{and}\ \ m\gamma_{\ell,n}  \xrightarrow{m \rightarrow \infty} \infty,
  \label{eq:cond-ratio}
\end{equation}
where $m=\min{(\ell,n)}$.

\section{\textit{V}-Matrices}
\label{sec:norm}

Let us rewrite the first term of functional
\eqref{eq:tikh-func-ratio}:
\begin{equation*}
\rho^2=\left\| \frac{1}{n}\sum_{i=1}^{n}{r(x'_i)\,\theta(x-x'_i)} -
\frac{1}{\ell}\sum_{i=1}^{\ell}{\theta(x-x_i)} \right\|_{E_2}^2.
\end{equation*}
Using the $L_2$ norm in space $E_2$, we obtain:
\begin{equation}
\rho^2=\bigintsss_{0}^{u}{\left[\frac{1}{n}\sum_{i=1}^{n}{r(x'_i)\,\theta(x-x'_i)}
- \frac{1}{\ell}\sum_{i=1}^{\ell}{\theta(x-x_i)}\right]^2\,dx}.
\label{eq:l2mu-norm-1}
\end{equation}
Expression \eqref{eq:l2mu-norm-1} can be written as
\begin{equation}
\rho^2=\int_{0}^{u}{a^2\,dx} -2 \int_{0}^{u}{ab\,dx} +
\int_{0}^{u}{b^2\,dx}, \label{eq:l2mu-norm-2}
\end{equation}
where
\begin{align*}
& a = \frac{1}{n}\sum_{i=1}^{n}{r(x'_i)\,\theta(x-x'_i)}, &  x'_1,\ldots,x'_n~\stackrel{i.i.d}{\sim}~F_2(x),
\\
& b =  \frac{1}{\ell}\sum_{i=1}^{\ell}{\theta(x-x_i)}, &  x_1,\ldots,x_\ell~\stackrel{i.i.d}{\sim}~F_1(x).
\end{align*}

The last term in \eqref{eq:l2mu-norm-2} does not depend on $r(x)$
and thus can be ignored for minimization on $r$. In what follows, we
use the notation
\begin{equation*}
(t_1 \vee t_2) = \text{max}(t_1,t_2).
\end{equation*}
The first two terms of \eqref{eq:l2mu-norm-2} are
\begin{equation*}
\begin{split}
& \int_{0}^{u}{a^2\,dx} = \\ & = \frac{1}{n^2}\sum_{i=1}^{n}{\sum_{j=1}^{n}{r(x'_i)\,r(x'_j)\int_{0}^{u}{\theta(x-x'_i)\,\theta(x-x'_j)}\,dx }} \\
& = \frac{1}{n^2}\sum_{i=1}^{n}{\sum_{j=1}^{n}{r(x'_i)\,r(x'_j)\int_{(x_i^{1'} \vee x_j^{1'})}^{u^1}\hspace{-1.5ex}\text{\scriptsize ...}\int_{(x_i^{d'} \vee x_j^{d'})}^{u^d}{\,dx^1\text{\scriptsize ...}\, dx^d} }}
\\ & = \frac{1}{n^2}\sum_{i=1}^{n}{\sum_{j=1}^{n}{r(x'_i)r(x'_j) \prod_{k=1}^{d}{\left[u^k - (x_i^{k'} \vee x_j^{k'})\right]} }}
\end{split}
\label{eq:term1}
\end{equation*}

\begin{equation*}
\begin{split}
& -2 \int_{0}^{u}{ab\,dx} = \\ & = -\frac{2}{n\ell}\sum_{i=1}^{n}\sum_{j=1}^{\ell}{r(x'_i)\int_{0}^{u}{\theta(x-x'_i)\,\theta(x-x_j)}\,dx } \\
& = -\frac{2}{n\ell}\sum_{i=1}^{n}\sum_{j=1}^{\ell}{r(x'_i)\int_{(x_i^{1'} \vee x_j^{1})}^{u^1}\ldots\int_{(x_i^{d'} \vee x_j^{d})}^{u^d}{\,dx^1 \ldots\, dx^d} }
\\ & = \frac{1}{n^2}\sum_{i=1}^{n}{\sum_{j=1}^{n}{r(x'_i)r(x'_j) \prod_{k=1}^{d}{\left[u^k - (x_i^{k'} \vee x_j^{k})\right]}}}
\end{split}
\label{eq:term2}
\end{equation*}

Let us denote by $v''_{ij}$ the values
\begin{align*}
& v''_{ij} = \prod_{k=1}^{d}{\left[u^k - (x_i^{k'} \vee x_j^{k'})\right]}, & x'_i,x'_j~\sim~F_2(x)
\end{align*}
and by $\mathbf{V''}$ the $n \times n$-dimensional matrix of
elements $v''_{ij}$. Also, let us denote by $v'_{ij}$ the values
\begin{align*}
& v'_{ij} = \prod_{k=1}^{d}{\left[u^k - (x_i^{k'} \vee x_j^{k})\right]}, & x'_i~\sim~F_2(x), x_j~\sim~F_1(x)
\end{align*}
and by $\mathbf{V'}$ the $n \times \ell$-dimensional matrix of
elements $v'_{ij}$.

Therefore, the first term of functional \eqref{eq:tikh-func-ratio} has the following form in vector-matrix notation:
\begin{equation}
\frac{1}{2} \vec{r}^\top\mathbf{V''}\vec{r} - \frac{\ell}{n} \vec{r}^\top\mathbf{V'}\vec{1} + \text{const},
\label{eq:norm-matrix}
\end{equation}
where by $\vec{r}$ we denote the $n$-dimensional vector
$\left[r(x'_1),\ldots,r(x'_n)\right]^\top$ and by $\vec{1}$ the
$\ell$-dimensional vector $\left[1,\ldots,1\right]^\top$.

Matrices $\mathbf{V''}$ and $\mathbf{V'}$ reflect the geometry
of the observed data.

\section{Regularizing Functional}
\label{sec:reg}

We define the regularizing functional $\Omega(r)$ to be the square of the norm in Hilbert space:
\begin{equation}
\Omega(r) = \left\|r\right\|^2_{\mathcal{H}} = \left\langle r,r \right\rangle_{\mathcal{H}}.
\label{eq:reg-hs}
\end{equation}
We will consider two different concepts of Hilbert space and its
norm, which correspond to two different types of prior knowledge
about the solution \footnote{Recall that the solution $r_0$ must
have a norm in the corresponding space: $\left\langle r_0,r_0
\right\rangle_{\mathcal{H}}=c_0<\infty$}.

Let us first define \eqref{eq:reg-hs} using the $L_2$ norm in Hilbert space:
\begin{equation}
\Omega(r) = \int{r(t)^2\,dF_{2,n}(t)} = \frac{1}{n} \sum_{i=1}^{n}{r(x'_i)^2} = \frac{1}{n} \mathbf{r^\top r}.
\label{eq:reg-1}
\end{equation}

We also define \eqref{eq:reg-hs} as a norm in a Reproducing Kernel
Hilbert Space~(RKHS).

\subsection{Reproducing Kernel Hilbert Space and Its Norm}

An RKHS is defined by a positive definite kernel function $k(x,y)$
and an inner product $\left\langle\, , \,
\right\rangle_{\mathcal{H}}$ for which the following reproducing
property holds true:
\begin{equation}
\left\langle f(x),k(x,y) \right\rangle_{\mathcal{H}} = f(y),\ \forall f \in \mathcal{H}.
\label{eq:repr-rkhs}
\end{equation}

Note that any positive definite function $k(x,y)$ has an expansion in terms of
its eigenvalues $\lambda_i \geq 0$ and eigenfunctions $\phi_i(x)$:
\begin{equation}
k(x,y) = \sum_{i=1}^{\infty}\lambda_i\phi_i(x)\phi_i(y).
\label{eq:exp-kernel}
\end{equation}

Let us consider the set of functions
\begin{equation}
f(x,c) = \sum_{i=1}^{\infty}c_i\phi_i(x)
\label{eq:set-rkhs}
\end{equation}
and the inner product
\begin{equation}
\left\langle f(x,c^*),f(x,c^{**}) \right\rangle_{\mathcal{H}} = \sum_{i=1}^{\infty}\frac{c_i^{*}c_i^{**}}{\lambda_i}.
\label{eq:inner-rkhs}
\end{equation}
Then, kernel~(\ref{eq:exp-kernel}), set of
functions~(\ref{eq:set-rkhs}), and inner
product~(\ref{eq:inner-rkhs}) define an RKHS. Indeed,
\begin{equation*}
\left\langle f(x),k(x,y) \right\rangle_{\mathcal{H}} = \left\langle \sum_{i=1}^{\infty}c_i\phi_i(x),
\sum_{i=1}^{\infty}\lambda_i\phi_i(x)\phi_i(y) \right\rangle_{\mathcal{H}} = \sum_{i=1}^{\infty}\frac{c_i\lambda_i\phi_i(y)}{\lambda_i} = f(y).
\end{equation*}

According to the representer theorem~\cite{Kimeldorf:70:ams}, the solution of \eqref{eq:tikh-func-ratio} has the form
\begin{align}
& r(x) = \sum_{i=1}^{n}\alpha_i\,k(x'_i,x), & x'_1,\ldots,x'_n \sim F_2(x).
\label{eq:sol-kernel}
\end{align}
Therefore, using \eqref{eq:repr-rkhs} and \eqref{eq:sol-kernel}, the
norm in RKHS has the form
\begin{equation}
\left\langle r(x),r(x) \right\rangle_{\mathcal{H}} =
\sum_{i,j=1}^{n}\alpha_i\alpha_j\,k(x'_i,x'_j) = {\vec\alpha}^\top
\mathbf{K} \vec{\alpha}, 
\label{eq:reg-2}
\end{equation}
where $\mathbf{K}$ denotes the $n \times n$-dimensional matrix with
elements $k_{ij} = k(x'_i,x'_j)$ and ${\vec\alpha}$ denotes the
$n$-dimensional vector $\left[\alpha_1,\ldots,\alpha_n\right]^\top$.
We choose this norm as a regularization functional:
$$
\Omega(r) = {\vec\alpha}^\top \mathbf{K} \vec{\alpha}.
$$

\section{Solving the Minimization Problem}
\label{sec:solving-min}


In this section we obtain solutions of minimization problem \eqref{eq:tikh-func-ratio}
for a fixed value of the regularization constant $\gamma$.

\subsection{Solution at Given Points (DRE-V)}

We rewrite minimization problem \eqref{eq:tikh-func-ratio} in an
explicit form using \eqref{eq:norm-matrix} and \eqref{eq:reg-1}.
This leads to the following optimization problem:
\begin{equation}
\mathcal{L} = \min_{\vec{r}} \left[\frac{1}{2}   \vec{r}^\top
\mathbf{V''} \vec{r} - \frac{n}{\ell}  \vec{r}^\top \mathbf{V'}\vec{1} +
\frac{\gamma}{n} \vec{r}^\top \vec{r}\right],
\label{eq:min-problem-1}
\end{equation}

The minimum of this functional has the form
\begin{equation}
\vec{r} = \frac{n}{\ell} (\mathbf{V''} +
\frac{\gamma}{n}\,\mathbf{I})^{-1}\mathbf{V'}\vec{1},
\label{eq:min-given-points}
\end{equation}
which can be computed by solving the corresponding system of linear equations.

In order to obtain a more accurate solution, we can take into account
our prior knowledge that
\begin{equation*}
r(x'_i) \geq 0, \ \ i=1,\ldots,n.
\end{equation*}
Any standard quadratic programming package can be used to find the
solution (vector $[r(x'_1),\ldots,r(x'_n)]$).

Optimization problem \eqref{eq:min-problem-1} has a structure
similar to that of the Kernel Mean Matching (KMM)
method~\cite{gretton:09:book}. The major difference is that our
method uses $V$-matrices, which describe the geometry of the
observed data.

\subsection{Solution in RKHS (DRE-VK)}

In order to ensure smoothness of the solution in
\eqref{eq:tikh-func-ratio}, we look for a function in an RKHS
defined by a kernel. According to the representer
theorem~\cite{Kimeldorf:70:ams}, the function has the form
\begin{equation}
r(x) = \sum_{i=1}^{n}\alpha_i\,k(x'_i,x).
\label{eq:sol-dre-vk}
\end{equation}
We rewrite the minimization problem \eqref{eq:tikh-func-ratio} in
explicit form using \eqref{eq:norm-matrix} and \eqref{eq:reg-2}.
Since $\vec{r} = \mathbf{K}\vec{\alpha}$, we obtain the following
optimization problem:
\begin{equation}
\mathcal{L} = \min_{\vec{\alpha} \in \mathbf{R}^n} \left[\frac{1}{2}
\vec{\alpha}^\top \mathbf{KV''K} \vec{\alpha} - \frac{n}{\ell}
\vec{\alpha}^\top \mathbf{KV'}\vec{1} + \gamma\,\vec{\alpha}^\top
\mathbf{K} \vec{\alpha}\right]. \label{eq:min-problem-2}
\end{equation}

The minimum of this functional has the form
\begin{equation}
\vec{\alpha} = \frac{n}{\ell} (\mathbf{V''K} +
\gamma\,\mathbf{I})^{-1}\mathbf{V'}\vec{1}, \label{eq:min-rkhs}
\end{equation}
which can also be computed by solving the corresponding system of linear equations.

Optimization problem \eqref{eq:min-problem-2} has a structure
similar to that of the unconstrained Least-Squares Importance
Filtering (uLSIF) method~\cite{kanamori:09:jmlr}, where, instead of
$\mathbf{V''}$ and $\mathbf{V'}$, one uses identity matrices $\mathbf{I}_{n \times n}$
and $\mathbf{I}_{n \times \ell}$, and, instead of regularization
functional $\vec{\alpha}^\top \mathbf{K} \vec{\alpha}$, one uses
$\vec{\alpha}^\top\vec{\alpha}$.

In order to use our prior knowledge about the non-negativity of the
solution, one can restrict the set of functions in
\eqref{eq:sol-dre-vk} with conditions
\begin{equation*}
\alpha_i \geq 0\ \ \text{and}\  \ k(x'_i,x) \geq 0,\ \ i = 1,\ldots,n.
\end{equation*}

\subsection{Linear INK-Splines Kernel}
\label{sec:ink-splines}

In what follows, we describe a kernel that generates linear splines with an
infinite number of
knots \cite{vapnik:98:book}. This smoothing kernel has good approximating
properties and has no free parameters.

We start by obtaining the kernel for functions defined on the interval $[0,u]$.
The multidimensional kernel for functions on $[0,u^1] \times \ldots \times [0,u^d]$
is the product of unidimensional kernels.

According to its definition, a $p$-th order spline with $m$ knots
\begin{equation*}
0 < t_1 < t_2 < \ldots < t_m \leq u
\end{equation*}
has the form
\begin{equation}
f(x) = \sum_{i=0}^{p} b^*_i x^i + \sum_{i=1}^{m}{b_i\,(x-t_i)_{+}^p},
\label{eq:splines-finite}
\end{equation}
where
\begin{equation*}
(x-t_i)_{+} = \left\{
  \begin{array}{l l}
    x-t_i, & \quad \text{if}\ x > t_i \\
    0, & \quad \text{otherwise.} \\
  \end{array} \right.
\end{equation*}
For $p=1$, expression (\ref{eq:splines-finite}) provides a piecewise
linear function, whereas for $p>1$, it provides a piecewise
polynomial function.

Now, let the number of knots $m \rightarrow \infty$.
Then~(\ref{eq:splines-finite}) becomes
\begin{equation}
f(x) = \sum_{i=0}^{p} b^*_i x^i + \int_{0}^{u}{b(t)\,(x-t)_{+}^p\,dt}.
\label{eq:splines-infinite}
\end{equation}

It is possible to consider function~(\ref{eq:splines-infinite}) as an
inner product in an RKHS:
\begin{equation*}
K_p(x,y) = \sum_{i=0}^{p} x^i y^i + \int_{0}^{u}{(x-t)_{+}^p\,(y-t)_{+}^p\,dt}.
\end{equation*}
For the case of linear splines with an infinite number of knots, \textit{i.e}, $p = 1$
in~(\ref{eq:splines-infinite}), this kernel has a simple closed form expression:
\begin{equation*}
K_1(x,y) = 1 + xy + \frac{1}{2}\left|x-y\right|(x \wedge y)^2 + \frac{1}{3}(x \wedge y)^3,
\label{eq:kernel-ink-splines}
\end{equation*}
where we denote $x \wedge y = \text{min}(x,y)$.

As mentioned, the multidimensional linear INK-spline is the coordinate-wise
product of linear INK-splines:
\begin{equation*}
K_{1,d}(x,y) = \prod_{k=1}^{d}{K_1(x^k,y^k)}.
\end{equation*}

\section{Selection of Regularization Constant}
\label{sec:gamma}

According to the results presented in
Section~\ref{sec:theory-ratio}, the regularization constant $\gamma$
should satisfy \eqref{eq:cond-ratio} for large $n$ and $\ell$.
For finite $n$ and $\ell$, we choose $\gamma$ as follows.

\subsection{Cross-Validation for DRE-VK}

For problem \eqref{eq:min-problem-2}, we choose the regularization
constant $\gamma$ using $k$-fold cross-validation based on the
minimization of the least-squares criterion~\cite{kanamori:09:jmlr}:
\begin{equation*}
J_0 = \frac{1}{2}\int{\left[r(x)-r_0(x)\right]^2\,p_2(x)\,dx} = \frac{1}{2}\int{r(x)^2\,p_2(x)\,dx} - \int{r(x)\,p_1(x)\,dx} + \text{const.}
\end{equation*}

We partition data sets
\begin{equation*}
\mathcal{X} = \{x_1,\ldots,x_\ell~\stackrel{i.i.d}{\sim}~F_1(x)\} \ \ \ \mbox{and} \ \ \ \mathcal{X}' = \{x'_1,\ldots,x'_n~\stackrel{i.i.d}{\sim}~F_2(x)\}
\end{equation*}
into $k$ disjoint sets
\begin{equation*}
\mathcal{Z}_i = \{z_{1,i},\ldots,z_{\ell/k,i}\} \ \ \ \mbox{and} \ \ \ \mathcal{Z}'_i = \{z'_{1,i},\ldots,z'_{n/k,i}\}.
\end{equation*}

Using samples $\mathcal{X} \backslash \mathcal{Z}_i$ and $\mathcal{X}' \backslash \mathcal{Z}'_i$,
a solution to the DRE-VK minimization problem
\begin{equation}
\vec{\alpha}_\gamma = \frac{n}{\ell} (\mathbf{V''}_i \mathbf{K}_i +
\gamma\,\mathbf{I})^{-1} \mathbf{V'}_i \vec{1} \label{min-cv}
\end{equation}
is constructed for constant $\gamma$. After that, the solution is evaluated by the empirical
least-squares criterion on samples $\mathcal{Z}_i$ and $\mathcal{Z}'_i$:
\begin{equation*}
J_{i}^{\gamma} = \frac{1}{2}\sum_{i=1}^{n/k}{r_\gamma(z'_i)^2} - \frac{n}{\ell} \sum_{i=1}^{\ell/k}{r_\gamma(z_i)}
\end{equation*}
This procedure is performed for each $i=1,\ldots,k$.

Then, from a set of regularization constant candidates $\Gamma =
\{\gamma_1,\ldots,\gamma_m\}$, we select $\gamma^{*}$ that minimizes
the least-squares criterion over all folds, \textit{i.e},
\begin{equation*}
\gamma^{*} = \argmin_{\gamma \in \Gamma} \sum_{i=1}^{k}{J_{i}^{\gamma}}
\end{equation*}

We obtain the final solution using the selected $\gamma^{*}$ and
unpartitioned samples $\mathcal{X}$,$\mathcal{X}'$:
\begin{equation}
\vec{\alpha} = \frac{n}{\ell} (\mathbf{V''K} +
\gamma^{*}\,\mathbf{I})^{-1}\mathbf{V'}\vec{1} \label{min-cv-all}
\end{equation}

\subsection{Cross-Validation for DRE-V}

The aforementioned procedure for $\gamma$ selection is not readily available for
estimation of values of the density ratio function at given points.
However, we take advantage of the fact that finding a minimum of
\begin{equation}
\mathcal{L} = \min_{\vec{\alpha} \in \mathbf{R}^n} \left[\frac{1}{2}
 \vec{\alpha}^\top \mathbf{V''V''} \vec{\alpha} - \frac{n}{\ell}
\vec{\alpha}^\top \mathbf{V'}\vec{1} + \frac{\gamma}{n}\,\vec{\alpha}^\top
\mathbf{V''} \vec{\alpha}\right] \label{min-given-points-1}
\end{equation}
leads to the same solution at given points as \eqref{eq:min-given-points} if the same value of $\gamma$
is used for both \eqref{eq:min-given-points} and \eqref{min-given-points-1}.

Indeed, the minimum of \eqref{min-given-points-1} is reached at
\begin{equation}
\vec{\alpha}_\gamma = \frac{n}{\ell} (\mathbf{V''V''} +
\gamma\,\mathbf{V''})^{-1}\mathbf{V'}\vec{1}. \label{min-given-points-2}
\end{equation}
Consequently, the solution at given points is:
\begin{equation*}
\vec{r} = \mathbf{V''}\vec{\alpha} = \frac{n}{\ell} \mathbf{V''}(\mathbf{V''V''} + \frac{\gamma}{n}\,\mathbf{V''})^{-1}\mathbf{V'}\vec{1} = \frac{n}{\ell} (\mathbf{V''} + \frac{\gamma}{n}\,\mathbf{I})^{-1}\mathbf{V'}\vec{1}.
\end{equation*}

In order to choose $\gamma$ for DRE-V, we use the same least-squares cross-validation
procedure described in the last section, but using \eqref{min-given-points-2}
instead of \eqref{min-cv} and \eqref{min-cv-all}.

\section{Experiments}
\label{sec:exp}

In this section we report experimental results for the methods of density ratio estimation introduced in
this paper: DRE-V (solution at given points) and DRE-VK (smooth solutions in RKHS).
For the latter we instantiate two versions: DRE-VK-INK, which uses the linear INK-splines
kernel described in Section~\ref{sec:ink-splines}; and DRE-VK-RBF, which
uses the Gaussian RBF kernel
\begin{equation*}
k(x,y) = \text{exp}\left(- \frac{\left\|x-y\right\|_2^2}{2\sigma^2}\right).
\end{equation*}
In all cases the regularization constant $\gamma$ is chosen by 5-fold cross-validation.
For DRE-VK-RBF, the extra ``smoothing'' parameter $\sigma^2$ is cross-validated along with $\gamma$.

For comparison purposes, we also run experiments for the Kernel Mean
Matching (KMM) procedure~\cite{gretton:09:book}, the Unconstrained
Least-Squares Importance Filtering (uLSIF)
algorithm~\cite{kanamori:09:jmlr}, and the Kullback-Leibler
Importance Estimation Procedure (KLIEP)~\cite{sugiyama:07:nips}. For
uLSIF and KLIEP, we use the code provided on the authors'
website\footnote{\url{http://sugiyama-www.cs.titech.ac.jp/~sugi/software/}},
leaving its parameters set to their default values with the
exception of the number of folds of uLSIF, which is set\footnote{In
our experiments, 5-fold uLSIF performed better than the default
leave-one-out uLSIF} to 5. For KMM, we follow the implementation
used in the experimental section of~\cite{gretton:09:book}. KLIEP,
uLSIF, and KMM use Gaussian RBF kernels. KLIEP and uLSIF select
$\sigma^2$ by cross-validation, whereas KMM estimates $\sigma^2$ by
the median distance between the input points.

We choose not to enforce the non-negativity of the solutions of DRE-V and DRE-VK-* in
order to conduct a fair comparison with respect to uLSIF. We note
that KMM and KLIEP do enforce
\begin{equation*}
r(x'_i) \geq 0, \ \ i=1,\ldots,n.
\end{equation*}

\subsection{Experimental Setting}

The experiments were conducted on synthetic data described in
Table~\ref{tb:models}. Models 4 and 6 were taken from the
experimental evaluation of previously proposed
methods~\cite{kanamori:09:jmlr}.


\begin{table}[!b]
\centering
\caption{Synthetic models}
\label{tb:models}

\begin{small}

\begin{tabular}{|@{\hskip 0.10in}c@{\hskip 0.15in}c@{\hskip 0.15in}|@{\hskip 0.15in}c@{\hskip 0.10in}c@{\hskip 0.15in}|@{\hskip 0.15in}c@{\hskip 0.10in}|}
\hline \hline  & & & & \\ [-1.5ex]
   Model \# &       Dim. &      $p_1(x)$ &      $p_2(x)$ &      Supp. \\
[0.5ex]
\hline  & & & & \\ [-1.5ex]
1 & 1 &       Beta &    Uniform &    $(0,1)$ \\

 &    &  $(0.5,0.5)$ &            &            \\
[0.5ex]
\hline  & & & & \\ [-1.5ex]
2 & 1 &       Beta &    Uniform &    $(0,1)$ \\

  &   &  $(2,2)$ &            &            \\
[0.5ex]
\hline  & & & & \\ [-1.5ex]
3 & 1 &       Beta &       Beta &    $(0,1)$ \\

 &  &  $(2,2)$ &  $(0.5,0.5)$ &            \\
[0.5ex]
\hline  & & & & \\ [-1.5ex]
4  & 1 &   Gaussian &   Gaussian & $(-\infty,+\infty)$ \\

   &   & $(2,1/4)$ & $(1,1/2)$ &            \\
[0.5ex]
\hline  & & & & \\ [-1.5ex]
5  & 1  &    Laplace &    Laplace & $(-\infty,+\infty)$ \\

  &   & $(2,1/4)$ & $(1,1/2)$ &            \\
[0.5ex]
\hline  & & & & \\ [-1.5ex]
6 & 20 &   Gaussian &   Gaussian & $(-\infty,+\infty)^{20}$ \\

 &    & $([1,0,...,0],I_{20})$ & $([0,0,...,0],I_{20})$ &  \\
[0.5ex]
\hline  & & & & \\ [-1.5ex]
7  & 20 &    Laplace &    Laplace & $(-\infty,+\infty)^{20}$ \\

   &    & $([1,0,...,0],I_{20})$ & $([0,0,...,0],I_{20})$ &  \\
[0.5ex]
\hline
\hline
\end{tabular}

\end{small}
\vskip -0.1in
\end{table}

For each model, we sample $\ell$ points from $p_1(x)$ and another
$n$ points from $p_2(x)$, with $\ell=n=m$ varying in
$\{50,100,200\}$ for unidimensional data and $\{100,200,500\}$ for
20-dimensional data. For each $m$, we perform 20 independent draws.

We evaluate the estimated density ratio at the points sampled from $p_2(x)$, since most
applications require the estimation of the density ratio only at those points. Accordingly,
we compare the estimate $\vec{r} = \left[r(x'_1),\ldots,r(x'_m)\right]^\top$ to the actual density ratio
\begin{equation*}
\vec{r_0} = \left[\frac{p_1(x'_1)}{p_2(x'_1)},\ldots,\frac{p_1(x'_m)}{p_2(x'_m)}\right]^\top
\end{equation*}
using the normalized root mean squared error (NRMSE):
\begin{equation*}
NRMSE(\vec{r},\vec{r_0}) = \frac{\left\| \vec{r}-\vec{r}_0
\right\|_2}{\left\| \vec{r_0} \right\|_2}.
\end{equation*}

\subsection{Results}

Experimental results are summarized in
Tables~\ref{tb:results-others} and \ref{tb:results-ours}. For each
$m$, we report the mean NRMSE and the standard deviation over the
independent draws.

\begin{table}[!b]
\setlength{\tabcolsep}{3pt}

\centering \caption{Mean NRMSE and standard deviation of previously
proposed methods (uLSIF, KLIEP, KMM) of density ratio estimation}
\label{tb:results-others}
\begin{small}

\begin{tabular}{|cr|@{\hskip 0.20in}cc@{\hskip 0.25in}cc@{\hskip 0.25in}cc@{\hskip 0.15in}|}
\hline
\hline & & & & & & & \\ [-1.5ex]

      M. \# &          $m$ & \multicolumn{ 2}{c@{\hskip 0.25in}}{uLSIF} & \multicolumn{ 2}{c@{\hskip 0.25in}}{KLIEP} & \multicolumn{ 2}{c@{\hskip 0.15in}|}{KMM} \\
\hline

\multicolumn{ 1}{|c}{1} &         50 &       0.74 &     (0.15) & {\bf 0.61} & {\bf (0.12)} &       1.40 &     (0.68) \\

\multicolumn{ 1}{|c}{} &        100 &       0.78 &     (0.12) & {\bf 0.62} & {\bf (0.18)} &       1.60 &     (1.10) \\

\multicolumn{ 1}{|c}{} &        200 &       0.72 &     (0.16) & {\bf 0.64} & {\bf (0.13)} &       0.89 &     (0.61) \\
\hline

\multicolumn{ 1}{|c}{2} &         50 &       0.50 &     (0.28) & {\bf 0.39} & {\bf (0.08)} &       0.98 &     (0.46) \\

\multicolumn{ 1}{|c}{} &        100 &       0.47 &     (0.27) & {\bf 0.32} & {\bf (0.11)} &       0.68 &     (0.32) \\

\multicolumn{ 1}{|c}{} &        200 & {\bf 0.27} & {\bf (0.19)} & {\bf 0.30} & {\bf (0.12)} &       0.36 &     (0.10) \\
\hline

\multicolumn{ 1}{|c}{3} &         50 &       0.78 &     (0.74) & {\bf 0.44} & {\bf (0.21)} &       1.10 &     (0.70) \\

\multicolumn{ 1}{|c}{} &        100 &       0.55 &     (0.23) & {\bf 0.33} & {\bf (0.19)} &       0.67 &     (0.39) \\

\multicolumn{ 1}{|c}{} &        200 &       0.32 &     (0.24) & {\bf 0.26} & {\bf (0.19)} &       0.31 &     (0.09) \\
\hline

\multicolumn{ 1}{|c}{4} &         50 &       2.90 &     (5.50) & {\bf 1.30} & {\bf (1.70)} &       2.00 &     (2.40) \\

\multicolumn{ 1}{|c}{} &        100 &       0.87 &     (0.48) & {\bf 0.55} & {\bf (0.27)} &       1.40 &     (0.60) \\

\multicolumn{ 1}{|c}{} &        200 &       0.61 &     (0.19) & {\bf 0.42} & {\bf (0.13)} &       2.00 &     (0.59) \\
\hline

\multicolumn{ 1}{|c}{5} &         50 &       0.77 &     (0.51) & {\bf 0.68} & {\bf (0.46)} &       1.50 &     (0.77) \\

\multicolumn{ 1}{|c}{} &        100 &       0.82 &     (0.24) & {\bf 0.41} & {\bf (0.23)} &       1.70 &     (0.42) \\

\multicolumn{ 1}{|c}{} &        200 &       0.55 &     (0.19) & {\bf 0.32} & {\bf (0.10)} &       2.00 &     (0.84) \\
\hline

\multicolumn{ 1}{|c}{6} &        100 & {\bf 0.76} & {\bf (0.06)} &       1.10 &     (1.60) &       0.85 &     (0.13) \\

\multicolumn{ 1}{|c}{} &        200 & {\bf 0.76} & {\bf (0.06)} &       1.20 &     (0.51) &       0.80 &     (0.07) \\

\multicolumn{ 1}{|c}{} &        500 & {\bf 0.75} & {\bf (0.04)} &       0.84 &     (0.23) &       0.89 &     (0.07) \\
\hline

\multicolumn{ 1}{|c}{7} &        100 & {\bf 0.68} & {\bf (0.03)} & {\bf 0.67} & {\bf (0.02)} &       0.83 &     (0.11) \\

\multicolumn{ 1}{|c}{} &        200 & {\bf 0.68} & {\bf (0.02)} & {\bf 0.67} & {\bf (0.02)} &       0.86 &     (0.07) \\

\multicolumn{ 1}{|c}{} &        500 & {\bf 0.67} & {\bf (0.01)} & {\bf 0.66} & {\bf (0.01)} &       0.93 &     (0.08) \\ [0.5ex]
\hline
\hline
\end{tabular}
\end{small}
\vskip -0.1in
\end{table}

Overall, the direct constructive methods of density ratio estimation
proposed in this paper achieve lower NRMSE than previously proposed
methods uLSIF, KLIEP, and KMM.

Among the methods proposed in this paper, the ones providing smooth
estimates in RKHS (DRE-VK-*)
perform better than DRE-V. It is worth noting that the use of linear
INK-splines kernel tends
to provide equally or better performing estimates than the ones
provided by the RBF kernel.



We believe that the advantage in accuracy of the methods proposed in
this paper is due to 1)~the information provided by the $V$-matrices
about the geometry of the data, and 2) the smoothness requirements
introduced by RKHS.

\begin{table}[!t]
\setlength{\tabcolsep}{3pt}

\centering \caption{Mean NRMSE and standard deviation of the methods
of density ratio estimation based on the $V$-matrix concept and comparison to the best results of Table~\ref{tb:results-others}}
\label{tb:results-ours}
\begin{small}

\begin{tabular}{|cr|@{\hskip 0.20in}cc@{\hskip 0.25in}cc@{\hskip 0.25in}cc@{\hskip 0.15in}|@{\hskip 0.15in}cc@{\hskip 0.15in}|}
\hline
\hline & & & & & & & & & \\ [-1.5ex]

      M. \# &          $m$ & \multicolumn{ 2}{c@{\hskip 0.25in}}{DRE-V} & \multicolumn{ 2}{c@{\hskip 0.25in}}{DRE-VK-INK} & \multicolumn{ 2}{c@{\hskip 0.15in}|@{\hskip 0.15in}}{DRE-VK-RBF} & \multicolumn{ 2}{c@{\hskip 0.15in}|}{Others' Best} \\
\hline

\multicolumn{ 1}{|c}{1} &         50 &       0.72 &     (0.19) & {\bf 0.59} & {\bf (0.15)} & {\bf 0.61} & {\bf (0.17)} & {\bf 0.61} & {\bf (0.12)} \\

\multicolumn{ 1}{|c}{} &        100 &       0.69 &     (0.18) & {\bf 0.57} & {\bf (0.20)} &       0.65 &     (0.23) &       0.62 &     (0.18) \\

\multicolumn{ 1}{|c}{} &        200 &       0.62 &     (0.15) & {\bf 0.52} & {\bf (0.14)} & {\bf 0.51} & {\bf (0.18)} &       0.64 &     (0.13) \\
\hline

\multicolumn{ 1}{|c}{2} &         50 & {\bf 0.27} & {\bf (0.09)} &       0.33 &     (0.15) & {\bf 0.24} & {\bf (0.11)} &       0.39 &     (0.08) \\

\multicolumn{ 1}{|c}{} &        100 & {\bf 0.27} & {\bf (0.13)} & {\bf 0.28} & {\bf (0.16)} & {\bf 0.27} & {\bf (0.18)} &       0.32 &     (0.11) \\

\multicolumn{ 1}{|c}{} &        200 & {\bf 0.18} & {\bf (0.05)} & {\bf 0.19} & {\bf (0.08)} & {\bf 0.19} & {\bf (0.10)} &       0.27 &     (0.19) \\
\hline

\multicolumn{ 1}{|c}{3} &         50 & {\bf 0.34} & {\bf (0.21)} & {\bf 0.34} & {\bf (0.15)} &       0.40 &     (0.30) &       0.44 &     (0.21) \\

\multicolumn{ 1}{|c}{} &        100 & {\bf 0.25} & {\bf (0.22)} & {\bf 0.22} & {\bf (0.11)} & {\bf 0.24} & {\bf (0.16)} &       0.33 &     (0.19) \\

\multicolumn{ 1}{|c}{} &        200 &       0.19 &     (0.11) & {\bf 0.15} & {\bf (0.07)} & {\bf 0.16} & {\bf (0.07)} &       0.26 &     (0.19) \\
\hline

\multicolumn{ 1}{|c}{4} &         50 &       1.20 &     (1.50) & {\bf 0.81} & {\bf (0.80)} &       0.90 &     (0.94) &       1.30 &     (1.70) \\

\multicolumn{ 1}{|c}{} &        100 &       0.63 &     (0.34) & {\bf 0.43} & {\bf (0.20)} &       0.54 &     (0.36) &       0.55 &     (0.27) \\

\multicolumn{ 1}{|c}{} &        200 &       0.45 &     (0.14) & {\bf 0.32} & {\bf (0.16)} &       0.43 &     (0.18) &       0.42 &     (0.13) \\
\hline

\multicolumn{ 1}{|c}{5} &         50 & {\bf 0.50} & {\bf (0.22)} &       0.65 &     (0.36) &       0.68 &     (0.50) &       0.68 &     (0.46) \\

\multicolumn{ 1}{|c}{} &        100 & {\bf 0.43} & {\bf (0.11)} &       0.55 &     (0.19) &       0.51 &     (0.31) & {\bf 0.41} & {\bf (0.23)} \\

\multicolumn{ 1}{|c}{} &        200 & {\bf 0.35} & {\bf (0.15)} &       0.41 &     (0.14) & {\bf 0.35} & {\bf (0.14)} & {\bf 0.32} & {\bf (0.10)} \\
\hline

\multicolumn{ 1}{|c}{6} &        100 &       0.83 &     (0.06) & {\bf 0.73} & {\bf (0.09)} & {\bf 0.73} & {\bf (0.09)} & {\bf 0.76} & {\bf (0.06)} \\

\multicolumn{ 1}{|c}{} &        200 &       0.79 &     (0.07) & {\bf 0.68} & {\bf (0.08)} & {\bf 0.67} & {\bf (0.08)} &       0.76 &     (0.06) \\

\multicolumn{ 1}{|c}{} &        500 &       0.62 &     (0.06) & {\bf 0.57} & {\bf (0.07)} & {\bf 0.57} & {\bf (0.07)} &       0.75 &     (0.04) \\
\hline

\multicolumn{ 1}{|c}{7} &        100 &       0.69 &     (0.06) & {\bf 0.60} & {\bf (0.04)} & {\bf 0.61} & {\bf (0.04)} &       0.67 &     (0.02) \\

\multicolumn{ 1}{|c}{} &        200 &       0.58 &     (0.06) & {\bf 0.54} & {\bf (0.05)} & {\bf 0.56} & {\bf (0.06)} &       0.67 &     (0.02) \\

\multicolumn{ 1}{|c}{} &        500 &       0.50 &     (0.03) & {\bf 0.42} & {\bf (0.03)} &       0.50 &     (0.03) &       0.66 &     (0.01) \\
[0.5ex]
\hline
\hline
\end{tabular}
\end{small}
\vskip -0.1in
\end{table}

\section{Conclusion}
\label{sec:final}

The direct constructive methods of density ratio estimation
presented in this paper can be used for solving many different
problems of applied statistics. These methods can be modified
depending on different definitions of norms in  space $E_2$ and
regularizing functionals in  space $E_1$. The $V$-matrices will
change accordingly, since its elements will be computed differently.


\subsubsection*{Acknowledgments}

Igor Braga is supported by the Sao Paulo Research Foundation - FAPESP.

\bibliographystyle{splncs}
\bibliography{refs}

\end{document}